\documentclass[runningheads]{llncs}
\usepackage[T1]{fontenc}
\usepackage{graphicx}
\usepackage[T1]{fontenc}
\usepackage[misc]{ifsym}

\usepackage{float}
\usepackage{comment}
\usepackage{graphicx}
\usepackage{color}
\usepackage{xcolor}
 \usepackage{multirow}
\usepackage{xspace}

\usepackage{adjustbox}
\usepackage{tikz}
\usetikzlibrary{shapes}
\usepackage{amsmath,amssymb}
\usepackage{booktabs}
\usepackage{todonotes}
\usepackage{caption}
\usepackage{xurl}
\usepackage[linesnumbered,ruled,vlined]{algorithm2e}

\def\N{\mathcal{N}\xspace} 

\def\M{\mathcal{M}\xspace} 
\def\im{{m}\xspace} 
\def\G{\mathcal{G}\xspace} 
\def\ig{{g}\xspace} 
\def\W{\mathcal{N}\xspace} 
\def\IP{\mathcal{V}\xspace} 

\def\gAsIP{\mathcal{B}\xspace}

\def\cover{{\it cover}\xspace} 

\def\freq{{\it freq}\xspace} 

\def\desc{{\it desc}\xspace}  

\def\lang{\mathcal{L}\xspace} 

\newcommand{\val}[2]{v_{\ig^{#1},\im^{#2}}}

 \newcommand{\NIP}[0]{NIP}
 \newcommand{\NIPC}[0]{NIP_{Q}}

\newcommand{\Lquerysampling}[1]{L_{Q}(#1)}
\newcommand{\Jquerysampling}[1]{J_{Q}(#1)}

 \newcommand{\PL}[4]{P^{L_{{#1}^{#2}_{#3}}}_{\val{}{}}(#4)}
 \newcommand{\PJ}[4]{P^{J_{{#1}^{#2}_{#3}}}_{\val{}{}}(#4)}

\newcommand{\INC}[3]{Inc^{#1}_{#2}(#3)}
\newcommand{\EXC}[3]{Exc^{#1}_{#2}(#3)}
\newcommand{\SUP}[3]{Sup^{#1}_{#2}(#3)}
\newcommand{\SUPEQ}[3]{SupEq^{#1}_{#2}(#3)}
\newcommand{\INF}[3]{Inf^{#1}_{#2}(#3)}
\newcommand{\INFEQ}[3]{InfEq^{#1}_{#2}(#3)}

\newcommand{\fips}[0]{\textsc{Fips}}

\newcommand{\Cfips}[0]{\textsc{CFips}}

\newcommand{\alea}[0]{\textsc{Uniform}}

\begin{document}
\title{Frequency-based Constrained Sampling for Interval Patterns}

%
\author{
Djawad Bekkoucha\inst{1, 2} \and
Abdelkader Ouali\inst{2} \and
Bruno Crémilleux\inst{2}
}

\authorrunning{D. Bekkoucha et al.}

\institute{
Laboratoire Interdisciplinaire des Sciences du Numérique (LISN),\\
Université Paris-Saclay, CNRS,\\
91405 Orsay, France\\
\email{firstname.lastname@lisn.fr}
\and
Université Caen Normandie, ENSICAEN, CNRS, Normandie Univ,\\
GREYC UMR6072, F-14000 Caen, France\\
\email{firstname.lastname@unicaen.fr}
}

\maketitle              
\begin{abstract}

Output space pattern sampling is a powerful alternative to exhaustive pattern mining for exploring large pattern spaces, as it enables users to focus on representative patterns drawn according to a chosen interestingness measure.
In this paper, we address the problem of sampling interval patterns under user-defined syntactic constraints. We introduce \Cfips{}, a sampling approach that incorporates constraints directly into the sampling procedure. The approach relies on a multi-step sampling framework and supports several syntactic constraints by decomposing them into elementary predicates on interval bounds while preserving exact sampling guarantees.
We formally prove that \Cfips{} samples interval patterns proportionally to their frequency within the constrained pattern space. The experimental results show that integrating constraints into the sampling procedure enables to complete mining tasks that would otherwise fail within a given time out.

\end{abstract}

\section{Introduction}

Data scientists have a central role for knowledge discovery from data. In practice, analysts want to interact (visualize, select, explore) not only with the data, but also with the patterns or models supported by the data. To carry out such processes, the field of pattern mining has shifted to user-centered methods~\cite{DBLP:series/lncs/Leeuwen14}. For that purpose, it is essential to produce high quality results within a very short time to ensure a tight coupling between the system and the analyst. Pattern sampling is one solution to this challenge~\cite{DBLP:journals/pvldb/alhassan2009,DBLP:conf/kdd/BoleyLPG11,DBLP:journals/datamine/DzyubaLR17}.

Pattern sampling aims to randomly select a pattern $\IP$ with probability proportional to its interestingness measure $m(\IP)$. For example, when using frequency as the interestingness measure, a pattern $\IP_1$ that is twice as frequent as a pattern $\IP_2$ is twice as likely to be sampled~\cite{DBLP:conf/kdd/BoleyLPG11}. Unfortunately, pattern sampling often tends to focus on parts of the search space with a high density of uninteresting patterns (i.e., many patterns with low values for $m$) because these subspaces contain many patterns. For example, in the case of sequential data and the frequency interestingness measure, there is a large number of long patterns with a low frequency. To minimize this phenomenon, known as the "long-tail phenomenon", Diop et al.~\cite{DBLP:journals/kais/DiopDGLS20} proposed adding a maximum length constraint to the sampling process. Still for frequency  measure but for Boolean data, Soulet~\cite{DBLP:conf/f-egc/Soulet23} defined a generic sampling method that incorporates a minimum frequency threshold to remove infrequent patterns. 

The idea of combining constraints and sampling seems natural. Indeed, there is no reason not to immediately remove from the search space any patterns that do not satisfy the constraints, i.e., patterns that do not satisfy the analyst's conditions. However, this is not necessarily simple. For example, even for the simple frequency constraint, it is necessary to consider not only a pattern but also its support, which makes the work more difficult. A naive method that consists of drawing a pattern and then rejecting it if it does not satisfy the constraint can lead to a large number of rejections. An exhaustive extraction of all patterns followed by post-processing does not scale and loses the advantages of the sampling. To the best of our knowledge, there are very few works combining sampling while pushing the constraint into the extraction process~\cite{DBLP:journals/kais/DiopDGLS20,DBLP:conf/pakdd/Diop22,DBLP:conf/f-egc/Soulet23}.

In this paper, we define \Cfips{} the first sampling approach that incorporates constraints directly into the sampling procedure for numerical data. We choose to keep the complete original information expressed by the numerical data by using \textit{Interval Patterns}~\cite{DBLP:conf/ijcai/KaytoueKN11}. This approach prevents the loss of information that occurs when data are converted into a binary representation~\cite{Dougherty95supervisedand}. \Cfips{} relies on a multi-step sampling framework~\cite{DBLP:conf/kdd/BoleyLPG11}. The key principle is $NIP_Q$, a constrained counting function that computes, for each object, the exact number of interval patterns covering it while satisfying the constraints, without explicit pattern enumeration. \Cfips{} handles queries made of conjunctions of syntactic constraints by decomposing them into elementary predicates on interval bounds.  We formally prove that \Cfips{} samples interval patterns proportionally to their frequency within the constrained pattern space. If no pattern satisfies the query, \Cfips{} instantly
indicates that there is no solution. The analyst does not have to wait for a process that could take a long time to end. The experimental results show the interest of integrating constraints into the sampling process. Among other things, this makes it possible to complete mining tasks that would otherwise fail within a given time out. 
Contrary to Boolean data, syntactic constraints with interval patterns cannot be handled by a preprocessing step~\cite{DBLP:conf/pkdd/BonchiGMP03a} because these constraints also affect the support for an interval pattern. Therefore, the task is harder for interval patterns.

The remainder of this paper is organized as follows. Section~\ref{sec:preleminaries} introduces the preliminaries and formalizes the constrained interval pattern sampling problem. Section~\ref{sec:rw} reviews related work on pattern sampling and their ability to incorporate constraints. Section~\ref{sec:CFIPS} presents \Cfips{}, our approach for sampling interval patterns under constraints, and its theoretical properties. Section~\ref{sec:exemple_contraintesAdmissibles} describes the syntactic constraints supported by the method.  Section~\ref{sec:experiments} reports the experimental evaluation. Finally, Section~\ref{sec:conclusion} concludes the paper and outlines future research directions.


\section{Preliminaries}
\label{sec:preleminaries}
\subsection{{Numerical Dataset}}
A {\it numerical dataset} $\N$ is defined by
a set of objects $\G$ where each object is described by a set of attributes $\M$.
Each attribute $m \in \M$ has a range $\W_m$ which {is a finite set containing all the values of the data occurring in attribute $m$}.
 An object $\ig \in  \G$ is defined by a vector of numerical values  $< \val{}{} >_{\forall \im \in\M}$.
A dataset where the values of all attributes are binary $\W_m=\{0,1\}, \forall m \in \M$, is a special case of a numerical dataset and referred as a {\it binary dataset}. 

\begin{example}
Table~\ref{dataset:exemple} shows a running example of a numerical dataset containing 5~objects $\G=\{g_1,g_2, g_3, g_4, g_5\}$, 
each object is described by 3~attributes $\M=\{m_1,m_2, m_3\}$.     
\end{example}

\begin{table}[]
	\centering
	\scalebox{1}{
	\begin{tabular}{c r r r}
		& $m_1$ & $m_2$ & $m_3$ \\
		\hline
		$g_1$&2 &8 &130\\
		$g_2$&4 &12 &102\\
		$g_3$&3 &7 &91\\
		$g_4$&2 &9 &101\\
		$g_5$&6 &12 &110\\
	\end{tabular}
	}
	\caption{A running example of a numerical dataset $\N$}
	\label{dataset:exemple}
\end{table}

\subsection{{Interval Patterns}}
Patterns in numerical datasets can be represented in many ways, we use the notion of Interval Pattern~\cite{DBLP:conf/ijcai/KaytoueKN11}
\label{def:intervalpatterns} to prevent the loss of information that occurs when data are converted into a binary representation~\cite{Dougherty95supervisedand}. An Interval Pattern is defined as a vector of intervals $\IP =  \langle [a_m, b_m] \rangle_{\forall m \in \M}$, where $a_m, b_m \in \W_m$ and $a_m \leq b_m$.
Each dimension of the vector $\IP$ corresponds to an attribute following a canonical order on the set of attributes $\M$. We denote $\gAsIP[g] =  \langle [\val{}{}, \val{}{}] \rangle_{ \forall \im \in \M}$ as the vector of intervals corresponding to an object identified by $\ig$. An object $\ig$ is an occurrence of the interval pattern $\IP$ if each interval in the vector $\gAsIP[g]$ is included in the interval of $\IP$, i.e. $\gAsIP[g] \sqsubseteq \IP \iff [\val{}{}, \val{}{}] \subseteq [a_m, b_m], \forall \im \in \M$.
The cover of $\IP$ in $\N$ is the set of objects $\ig \in \G$ occurring in $\IP$, i.e. {$\cover(\IP)= \{\ig \in \mathcal{G} ~|~   \gAsIP[g] \sqsubseteq \IP\}$.}
 \label{def:cover}
 
\begin{example}
  {In the example dataset of Table \ref{dataset:exemple}}, $\IP = \langle [3, 4], [7, 12], [91, 130] \rangle $ is an interval pattern covering the objects $\{g_2, g_3\}$. $\gAsIP[g_2]$$ =  $ $\langle [4, 4]$, $[12, 12],$ $[102,102] \rangle$ is the vector of intervals identified by the object $g_2$ and an occurrence of $\IP$.
\end{example}

\label{def:frequency}
The frequency of $\IP$ is the {cardinality} of its cover, i.e. $\freq(\IP)=|\cover(\IP)|$. 
Given a minimum frequency threshold $\theta$, the interval pattern $\IP$ is frequent if and only if $\freq(\IP) \geq \theta$.
\label{def:description} 
{The smallest description of a subset of objects $G \subseteq \G$ is the smallest interval pattern covering the set of objects $G$}. Formally the smallest description of $G$ is the interval pattern $\IP$ {such that for each} $\ig ~\in~ G$, $\ig$ is an occurrence of $\IP$, i.e. $\desc(G)= \langle[a_m,~b_m]\rangle_{ \forall \im \in \M} \text{ such that } a_m = min(\{\val{}{}~|~ g~ \in G \}) \text{ and } b_m = max(\{\val{}{}~|~ g~ \in G \})$.

Let $\lang_{\IP}$ be the language of interval patterns, which corresponds to the set of all possible interval patterns. The size of the search space is the product of the total number of possible intervals for each attribute. Formally this is given by:
\[\prod_{m \in \M}{\sum_{k=1}^{|\W_m|}{k}} = \prod_{m \in \M}{\frac{|\W_{m}| (|\W_{m}|+1)}{2}}\]

Consider the database presented in Table~\ref{dataset:exemple}. The total number of interval patterns in the search space is given by:

\[\underbrace{\frac{4 \times 5}{2} }_{m_1} \times \underbrace{\frac{4 \times 5}{2}}_{m_2} \times \underbrace{\frac{5 \times 6}{2}}_{m_3} = 1500 \text{ interval patterns}\]

This example illustrates the rapid growth of the interval pattern search space. In practice, users are interested in patterns satisfying specific constraints, which define a small subspace of interest. Exhaustive extraction followed by post-processing does not scale, and rejection-based sampling may lead to prohibitively high rejection rates. This motivates integrating constraints directly into pattern sampling procedure, enabling efficient exploration of the search space while ensuring that sampled patterns satisfy user-defined constraints.




\subsection{Problem Statement}
Let $\N$ be a numerical database and $\lang_{\IP}$ the language of interval patterns defined over $\N$. 
Let $q_1=\{\text{intervals of } m_1 \text{ containing a given value}\}$ and 
$q_2=\{\text{intervals of } m_2 \text{ with lower bound greater than } 12\}$ be example constraints. We denote by $Q=\{q_i \land \ldots \land q_n\}$ a query corresponding to a conjunction of $n$ constraints. We denote by $\lang_{\IP Q} \subseteq \lang_{\IP}$ the set of interval patterns satisfying  $Q$. Let $f$ be an interestingness measure defined over $\lang_{\IP}$.

The constrained interval pattern sampling problem consists in sampling 
$\IP_1, \ldots$, $ \IP_k \in \lang_{\IP Q}$ independently with replacement according to a probability distribution proportional to $f$, without enumerating the constrained pattern space. 
In this work, we focus on frequency-based sampling and aim to push syntactic constraints directly into the sampling procedure while preserving exact sampling guarantees and avoiding rejection-based post-processing.
\section{Related Works}
\label{sec:rw}

The literature on output-space pattern sampling can be broadly categorized into three families.

\textbf{Stochastic methods.} These methods rely on random sampling procedures such as Markov chain Monte Carlo (MCMC) algorithms. Alhassan et al.~\cite{DBLP:journals/pvldb/alhassan2009} introduced this family for sampling graph patterns proportionally to their frequency. Boley et al.~\cite{DBLP:conf/sdm/BoleyGG10} extended these ideas to itemsets, considering positive interestingness measures. Bendimerad et al.~\cite{DBLP:conf/ida/BendimeradLPRB20} sampled tiles proportionally to a subjective interest measure. Despite their accuracy, slow convergence remains a common limitation.

\textbf{Declarative methods.} In this family, combinatorial solvers are leveraged to sample patterns. Dzyuba et al.~\cite{DBLP:journals/datamine/DzyubaLR17} pioneered SAT-based sampling of itemsets over multiple interestingness measures. Extending these methods to other pattern languages requires adapting the encoding for each language, limiting their applicability.

\textbf{Multi-step methods.} Methods in this family perform sequential sampling by decomposing the interestingness measure across dataset objects. Boley et al.~\cite{DBLP:conf/kdd/BoleyLPG11} proposed a two-step procedure that first draws an object from the dataset, then samples a pattern covering it. Diop et al.~\cite{DBLP:journals/kais/DiopDGLS20} extended this framework to three-step sampling for sequential patterns under maximum length constraints. Soulet~\cite{DBLP:conf/f-egc/Soulet23} enforced minimum frequency constraints for itemsets, while Diop~\cite{DBLP:conf/pakdd/Diop22} sampled high average-utility itemsets under length constraints using a two-step procedure. Formal guarantees on the resulting sampling distribution are obtained when an appropriate decomposition of the interestingness measure is performed.

\textbf{Sampling patterns from numerical data.} Numerical pattern sampling approaches remain limited. Giacometti et al.~\cite{DBLP:conf/sdm/GiacomettiS18} sample neighborhood patterns in a continuous space according to a density measure. Bekkoucha et al.~\cite{BEKKOUCHA2026102566} introduced interval pattern sampling proportionally to measures such as frequency or frequency $\times$ hypervolume.

\textbf{Constraints in pattern sampling.} The presented methods differ in their capacity to handle constraints. Stochastic methods enforce constraints via rejection or penalty mechanisms, which can severely slow convergence when constraints are highly restrictive. Declarative approaches allow straightforward incorporation of constraints, but scalability issues arise due to exhaustive solvers. Multi-step methods only handle a few types of constraints, such as length or frequency constraints~\cite{DBLP:journals/kais/DiopDGLS20,DBLP:conf/pakdd/Diop22,DBLP:conf/f-egc/Soulet23}. Integrating other types requires problem-specific decompositions. To the best of our knowledge, existing approaches for numerical pattern sampling do not incorporate constraints directly into the sampling procedure. This motivates \Cfips{}, the first pattern sampling approach from numerical data to incorporate syntactic constraints while preserving exact sampling guarantees.

\section{Frequency-based Interval Pattern Sampling under Syntactic Constraints}
\label{sec:CFIPS}

This section introduces \Cfips{}, a method for sampling interval patterns proportionally to their frequency while satisfying a query $Q$, which is a conjunction of syntactic constraints.

\subsection{\Cfips{} Key Ideas}

A naive approach for sampling interval patterns satisfying a query $Q$ proportionally to their frequency would consist in first mining all interval patterns, filtering through post-processing those that do not satisfy $Q$, and finally drawing the remaining patterns according to their associated frequency. However, due to the size of the interval pattern search space, this approach is impractical.

Multi-step sampling procedures address the problem of sampling patterns according to a predefined probability distribution without enumerating the pattern space. These approaches sample patterns by traversing the database through its objects rather than exploring the entire pattern space. They rely on decomposing the chosen interestingness measure into a sum of utilities defined over the objects of the database \cite{DBLP:phd/hal/Diop20}. This decomposition enables a successive sampling procedure in which an object is first drawn according to its weight, and a pattern covering this object is then sampled according to its utility for this object. The combination of these successive steps then leads to samples that follow the targeted distribution. However, these approaches do not provide a direct way to incorporate constraints into the sampling procedure.

Our idea is to incorporate directly into the utility function the constraints of $Q$. In \Cfips{}, patterns violating at least one constraint are assigned zero utility and therefore cannot be sampled. Formally, for an interval pattern $\IP \in \lang_{\IP}$ and an object $g \in \N$, the constrained utility function is defined as:

\[
u(\IP, g, Q) =
\begin{cases}
1 & \text{if } g \in \cover_{\N}(\IP) \text{ and } \IP \text{ satisfies } Q, \\
0 & \text{otherwise.}
\end{cases}
\]

This formulation ensures that patterns violating at least one constraint in $Q$ contribute neither to object weights nor to the sampling probability. Consequently, constraints are enforced implicitly through the utility function, without requiring rejection-based sampling.


The frequency measure can be decomposed as a sum over the objects of the database~\cite{DBLP:conf/kdd/BoleyLPG11}. 
This decomposition relies on a counting function that computes the number of interval patterns covering a given object while satisfying the query $Q$. Building on this principle, we introduce $\NIP_Q$, a decomposition of the frequency measure into object utilities designed to handle constraints in a numerical setting.

\subsection{Counting Valid Interval Patterns Under Syntactic Constraints} 
\label{sec:NIPC}
To sample patterns that satisfy a query $Q$ composed of syntactic constraints proportionally to their frequency, we restrict the computation of frequency-based weights to valid patterns $\mathcal{L}_{\IP Q}$ (i.e. $ \sum_{\IP \in \mathcal{L}_{\IP Q}} \freq(\IP, \N)$). 
To avoid an explicit enumeration of this constrained pattern space, we follow the principle of the \textsc{NIP} function proposed in~\cite{BEKKOUCHA2026102566}. We define the counting function $NIP_Q : g \mapsto \mathbb{N}$, which directly incorporates constraints in the counting function and returns, for each object $g \in \G$, the exact number of interval patterns satisfying the constraints $Q$ and covering $g$.


To incorporate $c$ into $NIP_{Q}$, $c$ is decomposed into two predicates $P_{\val{}{}}^{L_c}$ and $P_{\val{}{}}^{J_c}$: the first filters candidate lower bounds $\Lquerysampling{\val{}{}}$, the second candidate upper bounds $\Jquerysampling{\val{}{}}$. Formally, $NIP_{Q}$ is defined as:

\begin{equation}
\label{eq:NIPC}
\NIPC(g) = \prod_{m \in \M} |\Lquerysampling{\val{}{}}| \cdot |\Jquerysampling{\val{}{}}|
\end{equation}

where:
\begin{itemize}
    \item $\Lquerysampling{\val{}{}} = \left\{ v \in \W_m \mid v \leq \val{}{}  \bigwedge_{i=1}^{|Q|} P_{\val{}{}}^{L_{c_i}}(v) \right\}$ denotes the set of admissible values according to $Q$ that can serve as lower bounds. Each predicate $P_{\val{}{}}^{L_{c_i}}(v)$ checks whether value $v$ satisfies constraint $c_i$ when considered as a lower bound.
    
    \item $\Jquerysampling{\val{}{}} = \left\{ v \in \W_m \mid v \geq \val{}{}  \bigwedge_{i=1}^{|Q|} P_{\val{}{}}^{J_{c_i}}(v) \right\}$ denotes the set of admissible values according to $Q$ that can serve as upper bounds. Each predicate $P_{\val{}{}}^{J_{c_i}}(v)$ checks whether value $v$ satisfies constraint $c_i$ when considered as an upper bound.
\end{itemize}

For each attribute $m$, Equation~\ref{eq:NIPC} evaluates the number of admissible lower bounds and upper bounds for an interval containing the value $\val{}{}$. The product of these two terms gives the number of valid intervals on $m$ containing $\val{}{}$ and satisfying $Q$. By multiplying the results across all attributes, $\NIPC(g)$ returns the total number of interval patterns covering $g$ while satisfying all syntactic constraints in $Q$.


A particularly interesting result is that $\NIPC$ identifies situations in which no pattern satisfies $Q$: if $\NIPC(g)=0$ for all $g \in \G$, then the constrained pattern space $\lang_{\IP Q}$ is empty. This property allows \Cfips{} to immediately report that no interval pattern satisfying $Q$ exists and thus the sampling can be stopped contrary  to rejection-based strategies that would endlessly try to draw patterns.

\begin{example}
Consider a numerical database $\N$, object $\ig_3 \in \G$, attribute $m_1 \in \M$ (see Table~\ref{dataset:exemple}), and a query $Q = \{Inc^{m_1}_{6}\}$ requiring that every pattern include the value $6$ in the interval associated with attribute $m_1$.

The value of $m_1$ for $\ig_3$ is $3$. The valid intervals containing $3$ and satisfying $Inc^{m_1}_{6}$ are $[2,6]$ and $[3,6]$, i.e., two intervals in total. Hence, for $m_1$:
\[
\Lquerysampling{3} = \{2,3\} \quad \text{and} \quad \Jquerysampling{3} = \{6\}
\]

These sets are obtained by incorporating the predicates associated with the constraint into each term:
\[
\begin{aligned}
\PL{Inc}{m_1}{6}{v} & : m = m_1 \land 6 \leq \val{}{} \Rightarrow v \leq 6 \\
\PJ{Inc}{m_1}{6}{v} & : m = m_1 \land 6 \geq \val{}{} \Rightarrow v \geq 6
\end{aligned}
\]

A value $v$ is a valid lower bound $\Lquerysampling{3}$ if it is less than or equal to the object's value $3$ for $m_1$ and satisfies $\PL{Inc}{m_1}{6}{v}$. Here, $6 \leq 3$ is false, so the predicate's premise does not apply, and the predicate is always true. The valid lower bounds are thus determined solely by $v \leq 3$ in $\W_{m_1}$, i.e., $\{2,3\}$.

For upper bounds, a value $v$ is admissible $\Jquerysampling{3}$ if it is greater than or equal to $3$ and satisfies $\PJ{Inc}{m_1}{6}{v}$. Since $6 \geq 3$ holds, the predicate enforces $v \geq 6$, restricting valid upper bounds to $\{6\}$.

Thus, $|\Lquerysampling{3}| \times |\Jquerysampling{3}| = 2 \times 1 = 2$.

Attributes $m_2$ and $m_3$ are unconstrained, admitting 4 and 5 valid intervals respectively. The total number of interval patterns covering $\ig_3$ and satisfying $Q$ is:
\[
\NIPC(\ig_3) = 2 \times 4 \times 5 = 40
\]
\end{example}

\subsection{\Cfips{} Sampling Algorithm}

We now describe the \Cfips{} sampling procedure (see Algorithm~\ref{algo:CFIPS}). The algorithm begins by determining the number of interval patterns satisfying the constraints in $Q$ that cover each object in the numerical database $\N$ (line~\ref{algocfips:0}). This step relies on the $\NIPC{}$ function (Section~\ref{sec:NIPC}). Line~\ref{algocfips:Step1} implements the first step of the two-step sampling procedure: an object $g$ is drawn with probability proportional to the number of interval patterns covering it and satisfying $Q$. The draw is therefore biased towards objects covered by a high number of patterns respecting the query constraints.

In the second step (line~\ref{algocfips:2}), an interval pattern satisfying all constraints in $Q$ and covering at least the selected object $g$ is drawn uniformly. Specifically, for each attribute $m \in \M$, two values $a_m$ and $b_m$ are drawn uniformly from $\Lquerysampling{\val{}{}}$ (admissible lower values) and $\Jquerysampling{\val{}{}}$ (admissible upper values), respectively. These values form the bounds of the interval associated with attribute $m$ in the sampled pattern.

A sample of $k$ patterns is obtained by executing $k$ times Algorithm~\ref{algo:CFIPS}. In the next section, we show that \Cfips{} draws patterns proportionally to their frequency.

\begin{algorithm}[h]
{
\textbf{Input:} numerical database $\N$; conjunction of syntactic constraints $Q$ \;
\textbf{Output:} a valid interval pattern $\IP^*$ satisfying $Q$, sampled proportionally to its frequency\;

\nl~\\

\textbf{Preprocessing:} compute $w_{F}(g) = \NIPC(g)$ for each object $g \in \G$ \label{algocfips:0}\;

\nl~\\

\textbf{Step 1:} $g \sim w_{F}(g)$ \quad \tcp*{Draw object $g$ proportionally to its weight $w_{F}$ \label{algocfips:Step1} }

\nl~\\
\textbf{Step 2:} \label{algocfips:2} \
\tcc*{Draw uniformly an interval pattern $\IP^*$ covering $g$ and satisfying $Q$}

$\IP^* \longleftarrow \langle \rangle$  \tcp*{Interval pattern under construction} 
\ForEach{attribute $m \in \M$}{
$a_m \longleftarrow$ uniformly draw a value from $\Lquerysampling{\val{}{}}$\;
$b_m \longleftarrow$ uniformly draw a value from $\Jquerysampling{\val{}{}}$\;
$\IP^* \longleftarrow \IP^* ++ [a_m,~ b_m] $\;
}

\nl~\\
\textbf{Return} $\IP^*$\;
}
\caption{Interval pattern sampling proportional to frequency under syntactic constraints (\Cfips{}) 
\label{algo:CFIPS}}
\end{algorithm}

\subsection{\Cfips{} Sampling Distribution}

We show that interval patterns produced by \Cfips{} are sampled proportionally to their frequency.

\begin{property}
\label{prop:Cfips:distribution}
For a numerical database $\N$, Algorithm~\ref{algo:CFIPS} draws each interval pattern $\IP$ with probability proportional to its frequency among patterns satisfying each constraint from the query $Q$.
\end{property}


\begin{proof}
\label{proof:distriCfips}
Let $\mathcal{L}_{\IP Q}\subseteq \mathcal{L}_{\IP}$ be the subspace of all interval patterns satisfying the user query $Q$, and let $Z = \sum_{\IP \in \mathcal{L}_{\IP Q}} |\cover(\IP, \N)|$ be the normalization constant, representing the sum of the frequencies of all interval patterns in $\mathcal{L}_{\IP Q}$. Let $g^{*} \in \G$ be an object drawn randomly in Step~1 of Algorithm~\ref{algo:CFIPS}, and let $\IP^{*}$ be an interval pattern drawn in Step~2 of the same algorithm. We then have:

\begin{align*}
P[\IP^{*} = \IP] &= \sum_{g \in \G} P[\IP^{*} = \IP \ \wedge \ g^{*} = g] \\
&= \sum_{g \in \cover(\IP, \N)} \frac{1}{\NIPC(g)} \cdot \frac{\NIPC(g)}{Z} \\
&= \sum_{g \in \cover(\IP, \N)} \frac{1}{Z} = \frac{|\cover(\IP, \N)|}{Z} = \frac{freq(\IP, \N)}{Z}
\end{align*}

where $Z = \sum_{g \in \G} \NIPC(g)$, which is equal to $Z = \sum_{\IP \in \mathcal{L}_{\IP Q}} |\cover(\IP, \N)|$, since each interval pattern contributes once for each object in its cover.
\end{proof}

\subsection{Time Complexity Analysis of \Cfips{}}

We analyze the time complexity of \Cfips{}, including the preprocessing phase and the sampling procedure.

\begin{property}
The overall time complexity of \Cfips{} for sampling a single interval pattern is:

\[
O(|\G|^2 \cdot |\M| \cdot \gamma + (\log|\G| + |\M|))
\]

where $\gamma$ is the worst-case complexity of evaluating the predicates associated with syntactic constraints over all distinct values.
\end{property}

\begin{proof}
Let $\N$ be a numerical database with $|\G|$ objects and $|\M|$ attributes. In the worst case, each attribute contains $|\G|$ distinct values, and each value participates in an unbounded number of constraints. The preprocessing complexity relies on the $\NIPC{}$ function applied to each object $g \in \G$. For each attribute $m \in \M$ and each distinct value $\val{}{}$ appearing in $g$, the sets $\Lquerysampling{\val{}{}}$ and $\Jquerysampling{\val{}{}}$ are constructed by filtering admissible bounds using the predicates defined by the syntactic constraints. Let $\gamma$ denote the worst-case complexity of evaluating all predicates for a value. The cost of constructing each set is $O(|\G|^2  \cdot |\M| \cdot \gamma)$.

Sampling a pattern with Algorithm~\ref{algo:CFIPS} is done in two steps. The first draws an object proportionally to its weight using a binary search of complexity $O(\log|\G|)$. The second draws a pattern covering this object uniformly from $\Lquerysampling{\val{}{}}$ and $\Jquerysampling{\val{}{}}$ in $O(|\M|)$.

Thus, the overall complexity for sampling a single pattern (including preprocessing) is:
\[
O(|\G|^2 \cdot |\M| \cdot \gamma + \log|\G| + |\M|).
\]
\end{proof}

To sample $k$ interval patterns, the preprocessing is performed once, and each draw costs $O(\log|\G| + |\M|)$. Therefore, the total complexity is:
\[
O(|\G|^2 \cdot |\M| \cdot \gamma + k(\log|\G| + |\M|)).
\]

\section{Examples of Admissible Constraints}
\label{sec:exemple_contraintesAdmissibles}

We illustrate examples of admissible syntactic constraints in \Cfips{}. These constraints can be expressed as independent predicates on the lower and upper bounds of interval patterns, making them directly integrable into the sampling procedure.

\begin{table}[h]
    \centering
    \renewcommand{\arraystretch}{1.5} 
    \begin{tabular}{|p{8cm}|p{6.5cm}|}
        \hline
        \textbf{Constraint Description} & \textbf{Formal Definition} \\
        \hline\hline

        \textbf{Inclusion:} ensures that a value $v' \in \W_m$ is included in the interval $[a_m, b_m]$ associated with attribute $m$. &
        $\INC{m}{v'}{\IP} \equiv a_m \leq v' \leq b_m$ with $a_m,~b_m \in \W_m$ \\
        \hline\hline

        \textbf{Exclusion:} ensures that a value $v' \in \W_m$ is excluded from the interval $[a_m, b_m]$ associated with attribute $m$. &
        $\EXC{m}{v'}{\IP} \equiv (v' < a_m) \lor (b_m < v')$ with $a_m,~b_m \in \W_m$ \\
        \hline\hline

        \textbf{Strictly greater than:} ensures that the interval $[a_m, b_m]$ associated with attribute $m$ is strictly greater than a value $v' \in \W_m$. &
        $\SUP{m}{v'}{\IP} \equiv v' < a_m \leq b_m$ with $a_m,~b_m \in \W_m$ \\
        \hline\hline

        \textbf{Greater than or equal:} ensures that the interval $[a_m, b_m]$ associated with attribute $m$ is greater than or equal to a value $v' \in \W_m$. &
        $\SUPEQ{m}{v'}{\IP} \equiv v' \leq a_m \leq b_m$ with $a_m,~b_m \in \W_m$ \\
        \hline\hline

        \textbf{Strictly less than:} ensures that the interval $[a_m, b_m]$ associated with attribute $m$ is strictly less than a value $v' \in \W_m$. &
        $\INF{m}{v'}{\IP} \equiv a_m \leq b_m < v'$ with $a_m,~b_m \in \W_m$ \\
        \hline\hline

        \textbf{Less than or equal:} ensures that the interval $[a_m, b_m]$ associated with attribute $m$ is less than or equal to a value $v' \in \W_m$. &
        $\INFEQ{m}{v'}{\IP} \equiv a_m \leq b_m \leq v'$ with $a_m,~b_m \in \W_m$ \\
        \hline
    \end{tabular}
    \caption{Examples of syntactic constraints defined over interval patterns}
    \label{tab:constraintList}
\end{table}

Table~\ref{tab:constraintList} provides a set of syntactic constraints that can be incorporated into the interval pattern sampling procedure. A user can specify a conjunction of these constraints in a query to sample patterns according to 
its interest.
To enable integration of these constraints in \Cfips{}, they are decomposed into elementary predicates applied to each bound, as detailed in Table~\ref{tab:predicateInstantiation}.

\begin{table}[h!]
\centering
\renewcommand{\arraystretch}{1.5} 
\begin{tabular}{|l|p{1.5cm} |p{6.2cm}|}
\hline
\textbf{Constraint} & \textbf{Term} & \textbf{Predicates} \\ \hline \hline
\multirow{2}{*}{Inclusion ($\INC{m'}{v'}{v}$)} 
& $\Lquerysampling{\val{}{}}$ & $\PL{Inc}{m'}{v'}{v}  : m = m' \land v' \leq \val{}{} \Rightarrow v \leq v'$ \\ 
 & $\Jquerysampling{\val{}{}}$ & $\PJ{Inc}{m'}{v'}{v}  : m = m' \land v' \geq \val{}{} \Rightarrow v \geq v'$ \\ 
\hline
\hline
\multirow{2}{*}{Exclusion ($\EXC{m'}{v'}{v}$)} 
 & $\Lquerysampling{\val{}{}}$ & $\PL{Exc}{m'}{v'}{v}  : m = m' \land v' \leq \val{}{} \Rightarrow v > v'$ \\ 
 & $\Jquerysampling{\val{}{}}$ & $\PJ{Exc}{m'}{v'}{v}  : m = m' \land v' \geq \val{}{} \Rightarrow v < v'$ \\ 
\hline
\hline
\multirow{2}{*}{Strictly greater than ($\SUP{m'}{v'}{v}$)} 
 & $\Lquerysampling{\val{}{}}$ & $\PL{Sup}{m'}{v'}{v}  : m = m' \Rightarrow v > v'$ \\ 
 & $\Jquerysampling{\val{}{}}$ & $\PJ{Sup}{m'}{v'}{v}  : m = m' \Rightarrow \top$ \\ 
\hline
\hline
\multirow{2}{*}{Greater or equal ($\SUPEQ{m'}{v'}{v}$)} 
 & $\Lquerysampling{\val{}{}}$ & $\PL{SupEQ}{m'}{v'}{v}  : m = m' \Rightarrow v \geq v'$  \\ 
 & $\Jquerysampling{\val{}{}}$ & $\PJ{SupEQ}{m'}{v'}{v}  : m = m' \Rightarrow \top$ \\ 
\hline
\hline
\multirow{2}{*}{Strictly less than ($\INF{m'}{v'}{v}$)} 
 & $\Lquerysampling{\val{}{}}$ & $\PJ{Inf}{m'}{v'}{v}  : m = m' \Rightarrow \top$ \\ 
 & $\Jquerysampling{\val{}{}}$ & $\PL{Inf}{m'}{v'}{v}  : m = m' \Rightarrow v < v'$ \\ 
\hline
\hline
\multirow{2}{*}{Less or equal ($\INFEQ{m'}{v'}{v}$)} 
 & $\Lquerysampling{\val{}{}}$ & $\PJ{InfEQ}{m'}{v'}{v}  : m = m' \Rightarrow \top$ \\ 
 & $\Jquerysampling{\val{}{}}$ & $\PL{InfEQ}{m'}{v'}{v}  : m = m' \Rightarrow v \leq v'$ \\ \hline 
 
\end{tabular}
\caption{Instantiation of predicates corresponding to the constraints presented in Table \ref{tab:constraintList}.
}
\label{tab:predicateInstantiation}
\end{table}


\textbf{Time complexity to evaluate each syntactic constraint.}  All predicates composing the constraints in Table \ref{tab:predicateInstantiation} can be evaluated in constant time, $\mathcal{O}(1)$, since they rely only on simple comparisons between the value $\val{}{}$, the distinct value under consideration $v \in \W_m$, and constants defined in the query ($m$ and $m'$). 
Therefore, the total cost to construct the sets $\Lquerysampling{\val{}{}}$ and $\Jquerysampling{\val{}{}}$ is linear in the number of distinct values, which, in the worst case, is $\mathcal{O}(|\G|)$.

\section{Experimental Evaluation}
\label{sec:experiments}
In this section, we evaluate the efficiency of \Cfips{} compared to classical sampling approaches followed by rejection-based post-processing to discard patterns that do not satisfy the constraints. We address the following resarch questions:

\begin{enumerate}
\item What is the rejection rate observed for classical sampling approaches combined with post-processing 
which rejects 
patterns that do not satisfy the constraints
to obtain the 
exact
number of valid patterns for a query $Q$?

\item In terms of computational time, how efficient is \Cfips{} compared to classical sampling approaches relying on post-processing to reject invalid patterns?
\end{enumerate}

\subsection{Experimental Protocol}

\textbf{Compared approaches.} We compare \Cfips{} with \fips{} and \alea{}. The two latter approaches were proposed in~\cite{BEKKOUCHA2026102566} for the interval pattern language. 
\fips{} samples interval patterns proportionally to their frequency, while \alea{} samples interval patterns uniformly. As both methods do not handle constraints, a post-processing step is required to check whether the sampled patterns satisfy the constraints specified in the query $Q$. This post-processing step consists in testing each sampled pattern against $Q$: if the pattern satisfies the constraints, it is retained; otherwise, it is rejected. The sampling process is repeated until the required number of valid patterns is obtained.

\noindent
\textbf{Datasets.} The experimental evaluation is conducted on five numerical datasets: \textit{cancer}, \textit{diabetes}, \textit{glass}, \textit{AP}, and \textit{NT}. Their characteristics are reported in Table~\ref{table:benchmark}.

\begin{table}[h]
    \centering
\scalebox{1}{
\begin{tabular}{|c||c|c|c|c|c|}\hline
    & NT & AP & Cancer & Glass & Diabetes \\\hline\hline
    $|\M|$ & 3 & 5 & 9 & 9 & 8 \\\hline
    $|\G|$ & 130 & 135 & 116 & 214 & 768 \\\hline
    $|Distinct\ values|$ & 67 & 674 & 900 & 939 & 1254 \\\hline
\end{tabular}}
\caption{Dataset characteristics $|\M|$ denotes the number of attributes, $|\G|$ the number of objects, and $|Distinct\ values|$ the total number of distinct values across all attributes.}
\label{table:benchmark}
\end{table}

\noindent
\textbf{Constraint generation.} For each dataset, a set of ten constraints is generated according to a protocol designed to simulate user behavior. An attribute $m$ is first drawn uniformly at random. If the number of distinct values $|\mathcal{W}_m|$ exceeds five, we assume that the user is interested in intervals located above or below a given threshold. In this case, a constraint is drawn uniformly from the set \{\textit{``greater than $v$'', ``less than $v$'', ``greater than or equal to $v$'', ``less than or equal to $v$''}\}, where $v \in \mathcal{W}_m$. 

If the attribute $m$ contains at most five distinct values, we assume that the user wishes to explicitly enforce or forbid the presence of certain values due to their semantic importance in the application domain. In this case, the constraint is chosen uniformly between \{\textit{``inclusion''}, \textit{``exclusion''}\}.

Finally, experiments are conducted by progressively adding constraints. The experiment starts with a single constraint; then a second constraint is added and the experiment is repeated with these two constraints. This process continues until all ten constraints are included. For each constraint set and each approach, 100 interval patterns are sampled. A time limit of two minutes is imposed for each run. Both computation time and rejection rate are averaged over ten repetitions. 

The source code, the datasets and the experimental results are available at \url{https://anonymous.4open.science/r/Constraint-Based-Sampling-6B83}

\subsection{Rejection Rate Evaluation}
\label{sec:exp_TauxRejet_CFIPS}

This section addresses Question~1, regarding the rejection rate observed for \fips{} and \alea{} methods. This rate corresponds to the proportion of sampled patterns that are invalid with respect to the query $Q$. Formally:
\[
\texttt{Rejection rate} = \left(1 - \frac{|K|}{\textit{NumberDraws}}\right) \times 100,
\]
where $|K|$ denotes the required number of patterns (fixed to $100$ in our experiments), and $\textit{NumberDraws}$ is the total number of draws needed to obtain $|K|$ patterns satisfying all constraints of the query $Q$.

\begin{figure}[htbp]
    \centering
    \begin{minipage}[b]{0.45\linewidth}
        \centering
        \includegraphics[width=\linewidth]{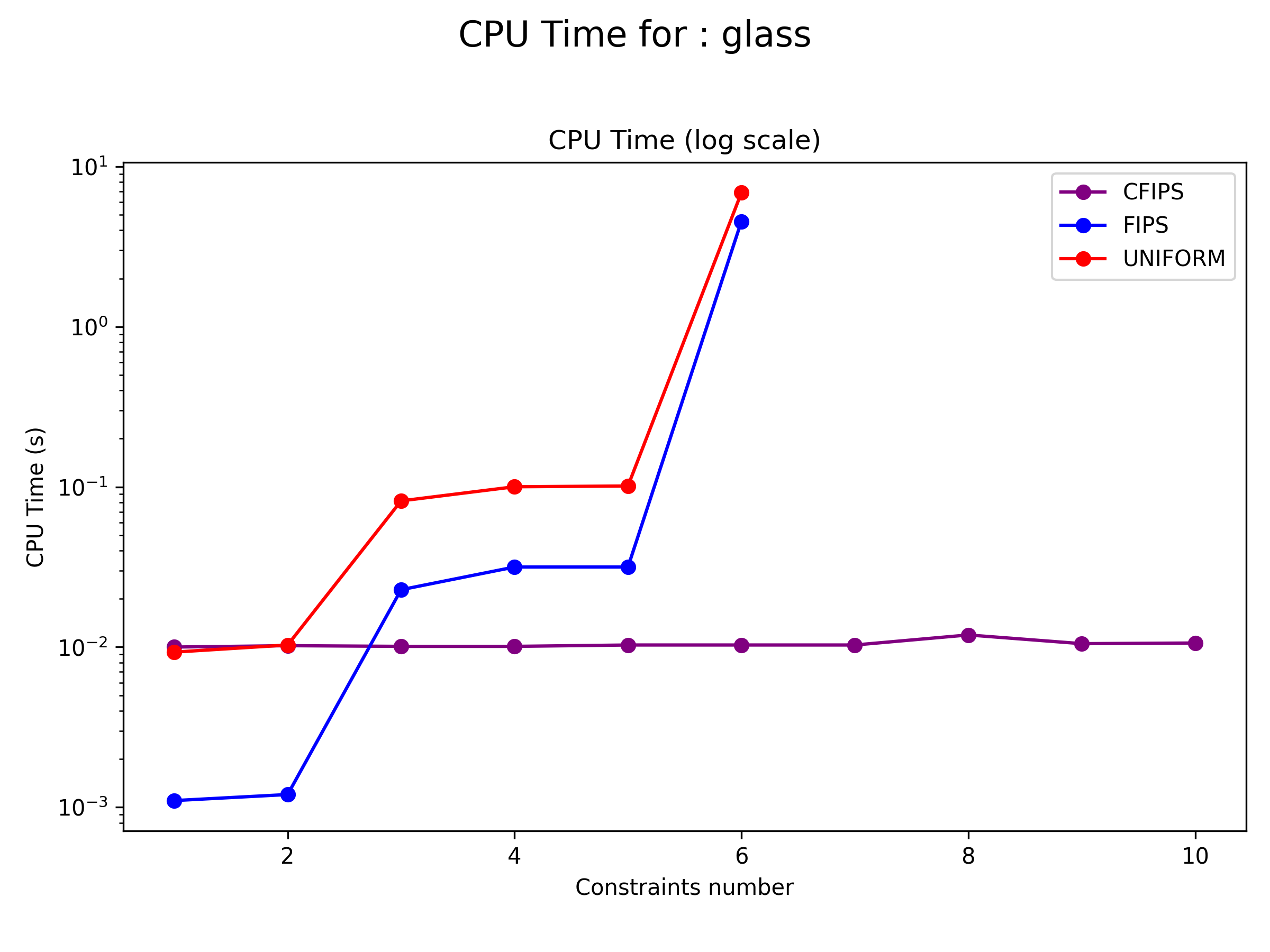}
    \end{minipage}
    \hspace{0.05\linewidth}
    \begin{minipage}[b]{0.45\linewidth}
        \centering
 {\footnotesize
    \setlength{\tabcolsep}{4pt}
    \renewcommand{\arraystretch}{0.9}
    \scalebox{0.8}{
    \begin{tabular}{|c||c|c|}
        \hline
        Constraints number & \fips{} & \alea{} \\\hline
        1 & 9.50 & 23.13 \\\hline
        2 & 17.89 & 28.05 \\\hline
        3 & 95.63 & 90.71 \\\hline
        4 & 96.79 & 92.59 \\\hline
        5 & 96.82 & 92.77 \\\hline
        6 & 99.97 & 99.89 \\\hline
        7 & TO & TO \\\hline
        8 & TO & TO \\\hline
        9 & TO & TO \\\hline
        10 & TO & TO \\\hline
    \end{tabular}
    }
    }
    \caption*{\small Rejection rate (\%) — \textit{glass}}
    \end{minipage}

    \vspace{0.5cm}

    \begin{minipage}[b]{0.45\linewidth}
        \centering
        \includegraphics[width=\linewidth]{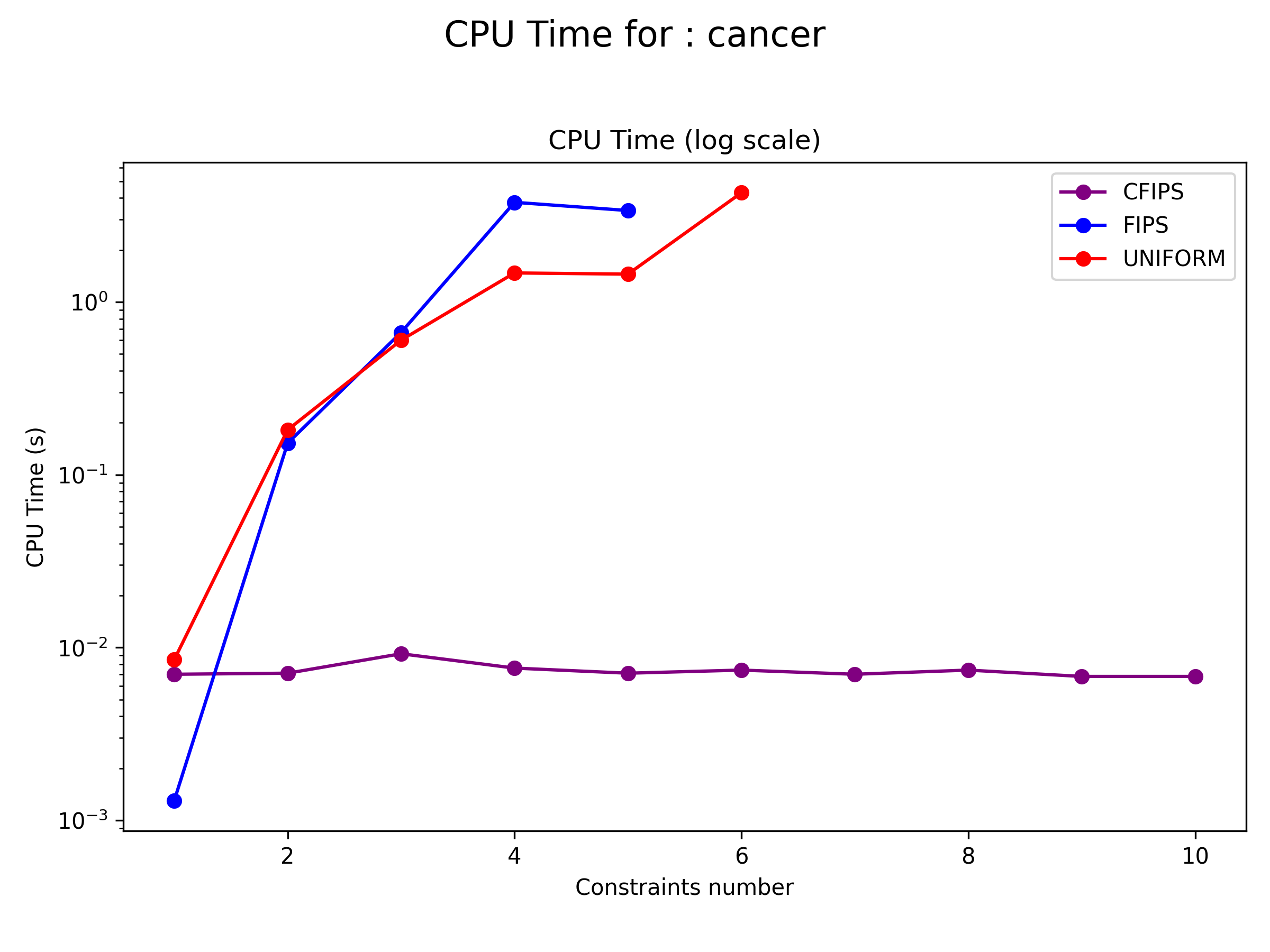}
    \end{minipage}
    \hspace{0.05\linewidth}
    \begin{minipage}[b]{0.45\linewidth}
        \centering
      {\footnotesize
    \setlength{\tabcolsep}{4pt}
    \renewcommand{\arraystretch}{0.9}
    \scalebox{0.8}{
    \begin{tabular}{|c||c|c|}
        \hline
        Constraints number & \fips{} & \alea{} \\\hline
        1 & 65.80 & 53.66 \\\hline
        2 & 99.69 & 97.96 \\\hline
        3 & 99.93 & 99.38 \\\hline
        4 & 99.98 & 99.75 \\\hline
        5 & 99.98 & 99.74 \\\hline
        6 & TO & 99.91 \\\hline
        7 & TO & TO \\\hline
        8 & TO & TO \\\hline
        9 & TO & TO \\\hline
        10 & TO & TO \\\hline
    \end{tabular}
    }
    }
    \caption*{\small Rejection rate (\%) — \textit{cancer}}
    \end{minipage}

    \vspace{0.5cm}

    \begin{minipage}[b]{0.45\linewidth}
        \centering
        \includegraphics[width=\linewidth]{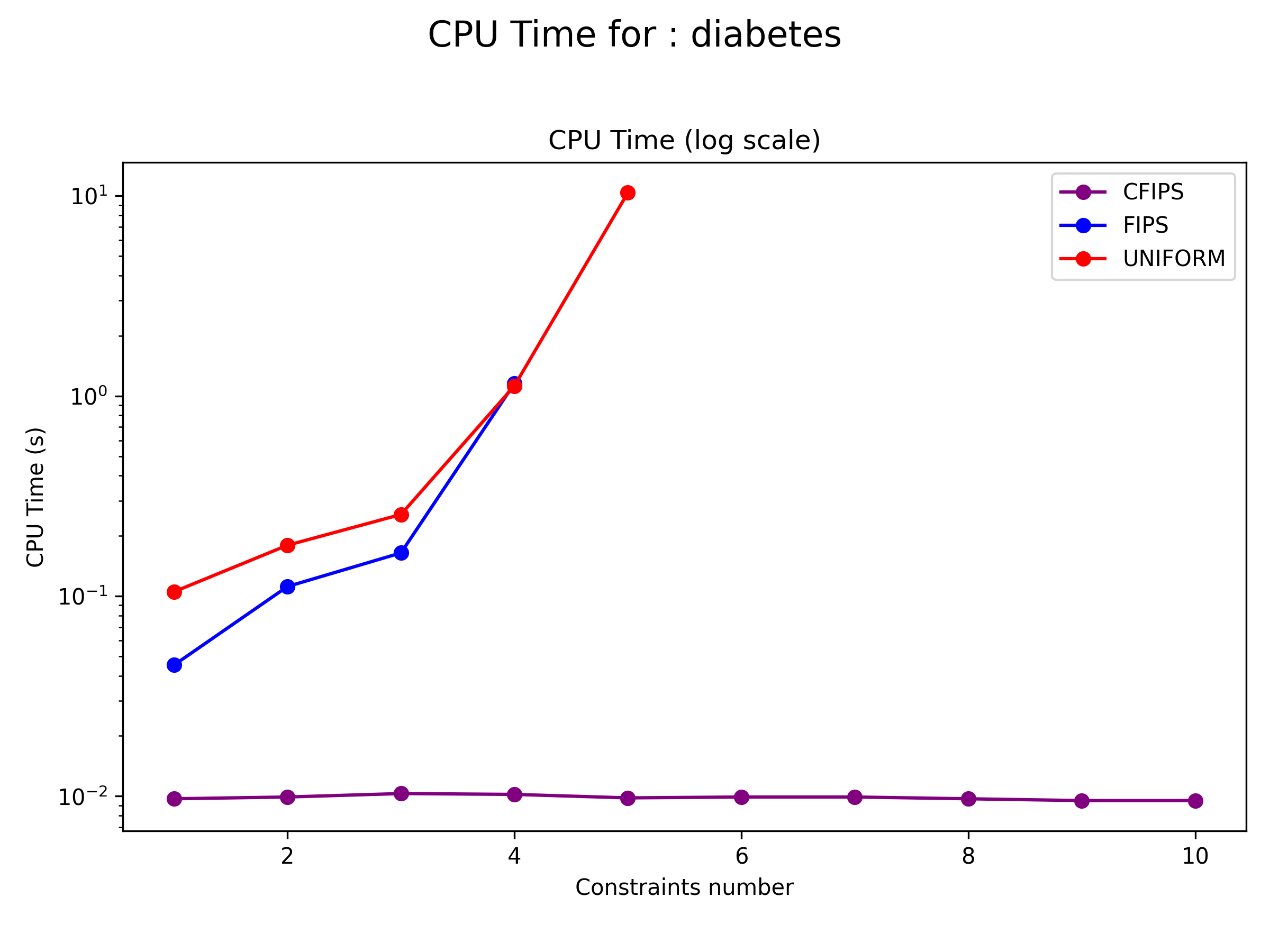}
    \end{minipage}
    \hspace{0.05\linewidth}
    \begin{minipage}[b]{0.45\linewidth}
        \centering
{\footnotesize
    \setlength{\tabcolsep}{4pt}
    \renewcommand{\arraystretch}{0.9}
    \scalebox{0.8}{
    \begin{tabular}{|c||c|c|}
        \hline
        Constraints number & \fips{} & \alea{} \\\hline
        1 & 95.03 & 85.72 \\\hline
        2 & 98.02 & 91.64 \\\hline
        3 & 98.63 & 93.96 \\\hline
        4 & 99.81 & 98.60 \\\hline
        5 & TO & 99.85 \\\hline
        6 & TO & TO \\\hline
        7 & TO & TO \\\hline
        8 & TO & TO \\\hline
        9 & TO & TO \\\hline
        10 & TO & TO \\\hline
    \end{tabular}
    }
    }
    \caption*{\small Rejection rate (\%) — \textit{diabetes}}
    \end{minipage}

    \vspace{0.5cm}

    \begin{minipage}[b]{0.45\linewidth}
        \centering
        \includegraphics[width=\linewidth]{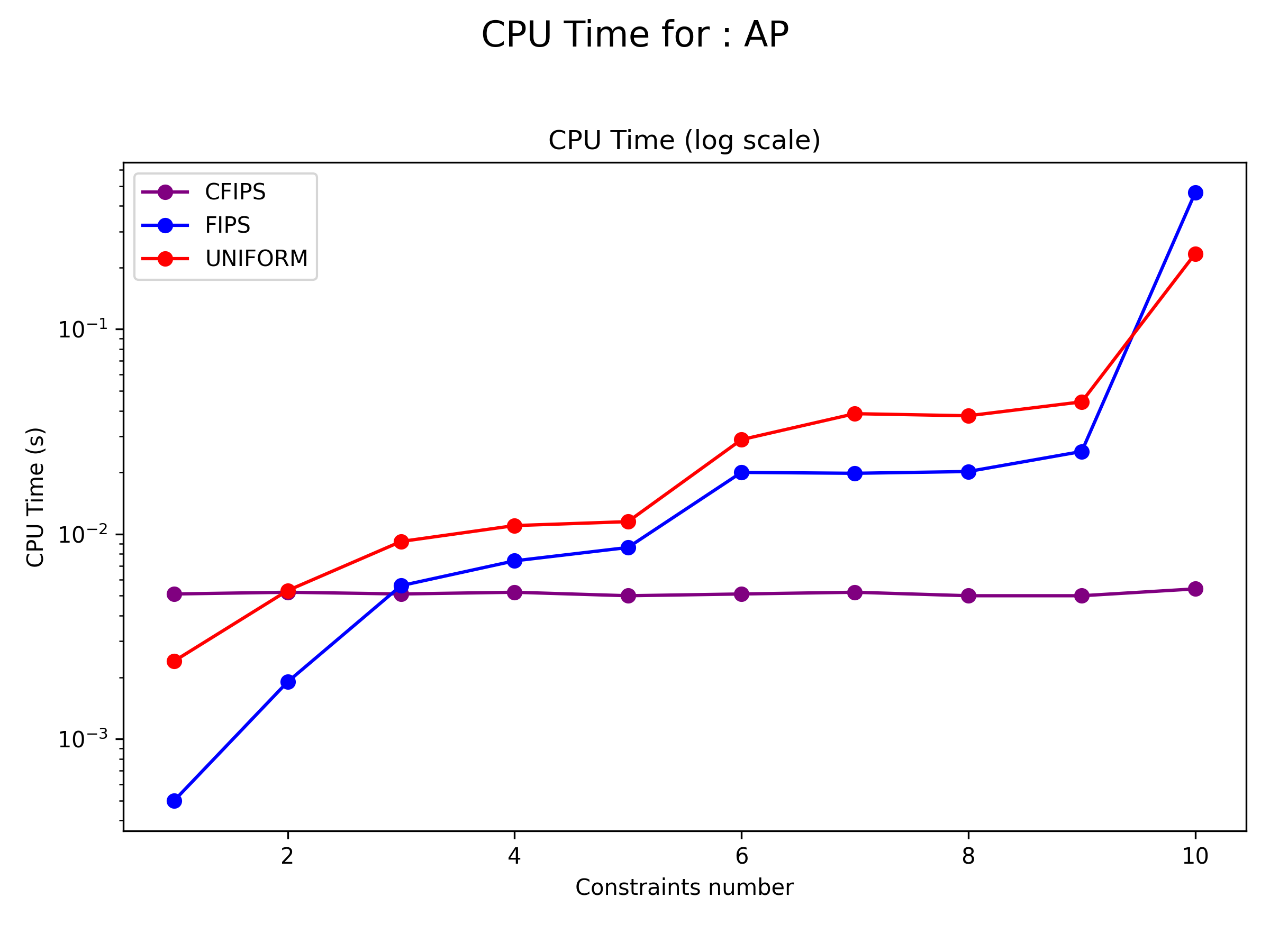}
    \end{minipage}
    \hspace{0.05\linewidth}
    \begin{minipage}[b]{0.45\linewidth}
        \centering
{\footnotesize
    \setlength{\tabcolsep}{4pt}
    \renewcommand{\arraystretch}{0.9}
    \scalebox{0.8}{
    \begin{tabular}{|c||c|c|}
        \hline
        Constraints number & \fips{} & \alea{} \\\hline
        1 & 0 & 0 \\\hline
        2 & 74.12 & 54.87 \\\hline
        3 & 89.21 & 73.62 \\\hline
        4 & 93.24 & 78.19 \\\hline
        5 & 93.90 & 79.18 \\\hline
        6 & 97.50 & 87.22 \\\hline
        7 & 97.51 & 87.60 \\\hline
        8 & 97.52 & 87.35 \\\hline
        9 & 98.05 & 89.11 \\\hline
        10 & 99.88 & 97.92 \\\hline
    \end{tabular}
    }
    }
    \caption*{\small Rejection rate (\%) — \textit{AP}}
    \end{minipage}

    \caption{Evaluation of CPU times and rejection rates for the \Cfips{}, \fips{}, and \alea{}}
    \label{fig:rejection_cpu_all}
\end{figure}

As presented on the right-hand side of Figure~\ref{fig:rejection_cpu_all}, 
\fips{} produces the required 100 patterns for up to 4 constraints on the \textit{diabetes} database, 5 on \textit{cancer}, and 6 on \textit{glass}. However, the rejection rates are high, particularly on \textit{cancer} and \textit{diabetes}, with values ranging from 65.80\% to 99.98\%. Beyond these thresholds, the method fails to generate the required number of patterns within the allocated time. An exception is observed for \textit{AP}, where \fips{} succeeds in sampling 100 patterns satisfying $Q$ even with up to 10 constraints. Nevertheless, the rejection rate still substantially increases with the number of constraints, rising from 74.12\% with 2 constraints to 99.88\% with 10. This can be explained by the interestingness measure of \fips{}, which tend to sample frequent patterns. This bias leads to producing patterns whose intervals contain a large number of distinct values, thus increasing the probability of including values that do not satisfy the constraints in $Q$.

\alea{} generally presents lower rejection rates than \fips{}, but still suffers from substantial rejection rates, from 23\% to 99\% on \textit{glass}, from 53\% to 99\% on \textit{cancer}, from 85\% to 99\% on \textit{diabetes}, and from 54\% to 97\% on \textit{AP}. For the latter database, \alea{} is able to generate the 100 required patterns regardless of the number of constraints. This behavior can be explained by the fact that Uniform tends to produce lower-frequency patterns \cite{BEKKOUCHA2026102566}, with smaller intervals that are therefore less likely to violate the constraints in $Q$.

\subsection{CPU Times Evaluation}

In this section, we address Question~2, which concerns the CPU time required to sample the desired number of patterns satisfying the constraints of the query $Q$. We compare \Cfips{} with \fips{} and \alea{}.

As presented on the left-hand side of Figure~\ref{fig:rejection_cpu_all}, for queries containing one to two constraints, the CPU times of \fips{} and \alea{} are sometimes lower than those of \Cfips{}, in particular on the \textit{AP} and \textit{glass} databases. However, as the number of constraints increases, the CPU time of \Cfips{} remains constant and becomes significantly lower than the CPU time of \fips{} and \alea{},  for which computation times grow substantially due to the increasing rejection rate.  From four or five constraints upward, \fips{} and \alea{} are no longer able to sample the required number of patterns within the allocated time, indicating a substantial increase in the number of rejected patterns. In an interactive context, these methods would not return solutions to an analyst within the allocated time.  Generally, \fips{} has the highest CPU times. This can be explained by its bias toward frequent patterns, which is more likely to violate the constraints in $Q$. This results in higher rejection rates and, consequently, longer computation times.

Note that interestingness measures affect the rejection rate in a constraint-dependent manner. \fips{}, which samples patterns according to frequency, produces wide intervals, which are more likely to include forbidden values and therefore violate exclusion-type constraints, while it can better satisfy inclusion-type constraints. Conversely, \alea{} samples patterns uniformly, producing narrow intervals across diverse regions of the search space \cite{BEKKOUCHA2026102566}; this reduces violations for exclusion constraints but may increase rejections for inclusion constraints, as it is less likely that the required values fall within the sampled intervals. By integrating constraints directly into the sampling procedure, \Cfips{} avoids these edge effects entirely.

\section{Conclusion and Perspectives}
\label{sec:conclusion}

In this paper, we presented \Cfips{}, the first constrained output-space interval pattern sampling approach from numerical data. \Cfips{} samples interval patterns proportional to their frequency while ensuring their validity with respect to a conjunction of constraints $Q$. Through $\NIP_Q$, we illustrated how constraints can be incorporated directly into the sampling procedure by decomposing them into independent predicates, ensuring that patterns are drawn according to the desired distribution in the constrained search space. Consequently, \Cfips{} can instantly indicate the absence of a solution, meaning the analyst does not have to wait. We theoretically proved that \Cfips{} samples patterns proportionally to their frequency in the constrained pattern space and experimentally demonstrated its advantages over post-processing rejection strategies. 

This work opens several research directions. A first direction is the integration of \Cfips{} into interactive pattern mining settings, where it could quickly return patterns matching an analyst’s preferences. Another direction is to extend the range of constraints supported by \Cfips{}. The current approach handles syntactic constraints that can be decomposed into elementary predicates on interval bounds. Extending it to more complex constraints, such as frequency thresholds or hypervolume, which cannot be easily decomposed, remains an open challenge. Finally, investigating additional interestingness measures for numerical data, such as density, constitutes another promising direction.

\bibliographystyle{elsarticle-num}
\bibliography{biblio}
\end{document}